\documentclass[a4paper,conference]{IEEEtran}
\usepackage{ifpdf}

\usepackage{cite}

\usepackage{graphicx}
\usepackage{xcolor}

\ifCLASSINFOpdf
\else
\fi
\usepackage{amsmath}
\usepackage{amssymb}
\usepackage{algorithmic}

\usepackage{array}

\ifCLASSOPTIONcompsoc
 \usepackage[caption=false,font=normalsize,labelfont=sf,textfont=sf]{subfig}
\else
 \usepackage[caption=false,font=footnotesize]{subfig}
\fi

\usepackage{stfloats}
\usepackage{url}

\begin{document}
%
\title{On the Evaluation of Generative Adversarial Networks By Discriminative Models}

\author{\IEEEauthorblockN{Amirsina Torfi}
\IEEEauthorblockA{Department of Computer Science\\
Virginia Tech\\
Blacksburg, VA 24061\\
Email: amirsina.torfi@gmail.com}
\and
\IEEEauthorblockN{Mohammadreza Beyki}
\IEEEauthorblockA{Department of Computer Science\\
Virginia Tech\\
Blacksburg, VA 24061\\
Email: mohi.beyki@gmail.com}
\and
\IEEEauthorblockN{Edward A. Fox}
\IEEEauthorblockA{Department of Computer Science\\
Virginia Tech\\
Blacksburg, VA 24061\\
Email: fox@vt.edu}}


%


\maketitle

\begin{abstract}
Generative Adversarial Networks (GANs) can accurately model complex multi-dimensional data and generate realistic samples. However, due to their implicit estimation of data distributions, their evaluation is a challenging task. The majority of research efforts associated with tackling this issue were validated by qualitative visual evaluation. Such approaches do not generalize well beyond the image domain. Since many of those evaluation metrics are proposed and bound to the vision domain, they are difficult to apply to other domains. Quantitative measures are necessary to better guide the training and comparison of different GANs models. In this work, we leverage Siamese neural networks to propose a domain-agnostic evaluation metric:  (1) with a qualitative evaluation that is consistent with human evaluation, (2) that is robust relative to common GAN issues such as mode dropping and invention, and (3) does not require any pretrained classifier. The empirical results in this paper demonstrate the superiority of this method compared to the popular Inception Score and are competitive with the FID score. 
\end{abstract}


%
\IEEEpeerreviewmaketitle

\section{Introduction}

\begin{figure*}[ht]
\centering
\includegraphics[width=0.8\textwidth]{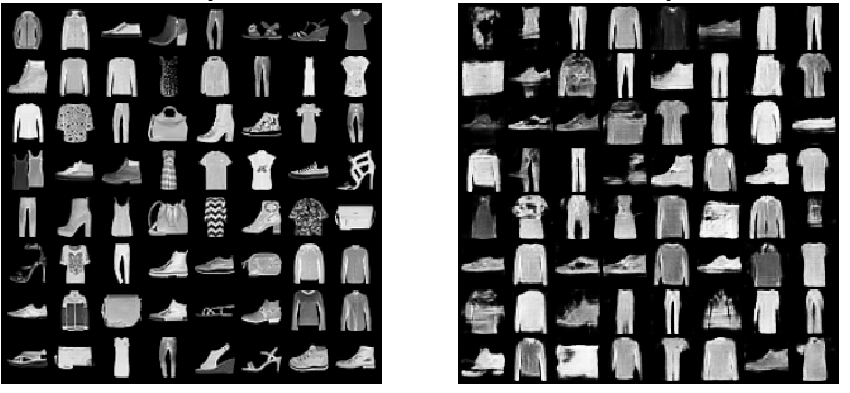}
\caption{Comparison of real~(left figure) and fake images~(right figure).~The fake images are generated using WGAN~\cite{arjovsky2017wasserstein} for training regime and DCGAN~\cite{radford2015unsupervised} as the architecture.}
\label{fig:fashiongan}
\end{figure*}

Generative Adversarial Networks (GANs) \cite{goodfellow2014generative} have gained much attention due to their capability to capture data characteristics, producing fake but realistic samples~(Figure~\ref{fig:fashiongan}),~and their superiority compared to other generative models.~Many successful research efforts utilized GANs in different applications such as image super-resolution~\cite{ledig2017photo},~ natural language generation~\cite{guo2018long},~healthcare synthetic data generation~\cite{esteban2017real},~style transfer~\cite{zhu2017unpaired}, etc.~\cite{karras2017progressive,mirza2014conditional,karras2019style,isola2017image}.

Despite GANs success for generative purposes, a significant challenge is their quantitative evaluation.~This is because the estimation of the underlying data distribution by a GAN model is only implicit.~We can generate fake samples solely based on the pre-trained distribution without the ability to calculate the intractable likelihood.~Even if we can do so, research shows that it can be misleading for high-dimensional complex data~\cite{theis2015note}. 

Many different GANs evaluation approaches have been proposed.
The majority of research efforts conducted associated with the image domain utilize visual evaluation of the synthesized sample.~Such visual approaches are subjective and might even be misleading~\cite{gerhard2013sensitive,sajjadi2018assessing}.~Furthermore, in other domains such as Natural Language Processing, it is not always straightforward or even plausible to visually evaluate the fake data.
Indeed, the generalizability of GANs evaluation metrics is a challenging topic of discussion, and only a few research efforts have been conducted regarding domain agnostic evaluation metrics~\cite{grnarova2019domain,sajjadi2018assessing}.

\begin{figure}[ht]
\centering
\includegraphics[width=0.35\textwidth]{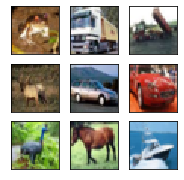}
\caption{The prediction of Inception Score on images from Cifar-10~(black) and the real class labels~(\textcolor{red}{red}).~Clearly IS score fails to go beyond ImageNet.~From left to right and top to bottom:~Amphibian~(\textcolor{red}{Frog}),~Milk Can~(\textcolor{red}{Truck}),~Milk Can~(\textcolor{red}{Truck}),~Threshing Machine~(\textcolor{red}{Deer}),~Sorrel~(\textcolor{red}{Automobile}),~Sorrel~(\textcolor{red}{Automobile}),~Container Ship~(\textcolor{red}{Bird}),~Japanese Spaniel~(\textcolor{red}{Horse}),~Fox Squirrel~(\textcolor{red}{Ship})}.
\label{fig:cifar10inception}
\end{figure}

Efforts in this domain led to the invention of metrics such as \textit{Inception Score (IS)}~\cite{salimans2016improved} and \textit{Fréchet Inception Distance (FID)}~\cite{heusel2017gans} that demonstrated promising evaluation results.~However, IS is proven to have many weaknesses such as suboptimality,~issues regarding its application beyond the ImageNet dataset~\cite{deng2009imagenet}~(see Figure~\ref{fig:cifar10inception}), and not being sensitive to mode dropping~\cite{barratt2018note}.~The FID score remedies the majority of IS issues, however, it is still not designed to operate beyond the vision domain without modification.

Motivated by such problems, we propose a novel approach to evaluate the GANs using \textit{Siamese Neural Networks~(SNNs)}.
These are known to be effective discriminative models for verification applications.~SNNs were previously used to augment the training of GANs~\cite{donahue2017semantically}.~In this work, we leverage SNNs to quantify the evaluation of GANs.

\textbf{Our Contributions:}
We \textbf{(1)}~introduce a novel metric to evaluate the GANs, demonstrating how it works and what desired characteristics it has;~\textbf{(2)}~compare our method with two other widely used approaches, to clarify our model's advantages;~and \textbf{(3)}~demonstrate that our method is domain agnostic by going beyond the vision domain.

\section{Related Works}~\label{sec:related}

Next, we discuss some of the well-known GANs evaluation methods and their advantages and disadvantages.~Many different approaches were proposed so far for this aim~\cite{wu2016quantitative,olsson2018skill,azadi2018discriminator}
One of the most popular metrics to evaluate the quality of generated images is called \textit{Inception Score}~(IS)~\cite{salimans2016improved}.
Although the metric was confirmed to relate well with the human evaluation of the naturalness of generated images,~it has proven to be inefficient and misleading in its applications~\cite{barratt2018note}.
One particular issue with IS is that it can only be used in the image domain, not in other domains such as text.
Fréchet Inception Distance~(FID) score~\cite{heusel2017gans} is an alternative metric proposed that remedies some of the setbacks of IS.
FID can identify intra-class mode dropping\footnote{Intra-class mode dropping refers to the situation that GAN only generates one or a few samples of each specific class.} which is a drawback in IS.

There are other evaluation metrics such as \textit{AIS}\cite{wu2016quantitative}, \textit{Skill rating} \cite{olsson2018skill,azadi2018discriminator}, and \textit{Precision-Recall} \cite{sajjadi2018assessing}.~AIS tries to estimate the log-likelihood using an assumed Gaussian observation which has been proven to have inaccurate estimates~\cite{grover2017flow}.~Skill rating utilizes the discriminatory information for evaluation, and the Precision-Recall approach is proposed to distinguish between different GANs failure cases. 

Beyond the utilization of metrics for evaluation of GANs,~a widely used approach is to train a classifier on fake samples and use a test set extracted from the real data to assess the data fidelity~\cite{im2018quantitatively,lopez2016revisiting}.~Although practical, such an approach limits the evaluation to the use-case under investigation and neither the classifier nor the training regime can be generalized to other use-cases. 

The majority of evaluation metrics are domain-specific,~fail to address mode dropping, or have drawbacks in terms of generalization.
For example, when  comparing our method with FID, although our approach does need labeled data, it does not need a pretrained classifier as is required in FID.~Furthermore, as FID operates on a pretrained image classifier,~its generalizability to other domains is problematic.~Accordingly, \textit{we propose a domain-agnostic metric to evaluate GANs} by the utilization of SNNs, and provide a comprehensive analysis.



    
    
    

\section{Proposed Approach}

\textit{How can we train a discriminative model to measure the realistic characteristics of fake data?}~We propose to  \textbf{\textit{train a discriminator using the real data}}.~This discriminator is independent of the discriminator trained by the GAN; we call it \textit{Siamese discriminator}.~For this aim,~we employ Siamese Neural Networks.

As a human observes a new pattern, he/she usually has the ability to recognize and associate this pattern with an already known concept, with a reasonable level of confidence~(Figure~\ref{fig:similarity}).
For evaluating the GANs, this is a challenging task, as not much supervised information is available about the generated data, since neither the label nor the explicit prior distributions are available.
However, given the discriminative model paradigm,~we aim to leverage the implicit distribution of the real data $p_X(x)$ to distinguish the realistic characteristics of the generated data $p_g(x)$.~By realistic we refer to the similarity of $p_g(x)$ to $p_X(x)$.
Thus, the goal here is to quantify this realism.

We train a discriminative model via a supervised learning paradigm with Siamese neural networks,~then we reuse the trained model to measure how well the GAN works. One important aspect of this quality assessment is the quality measurement of the generated samples. The main focus here is the character and image recognition. \textit{However, this approach is not bound to the domain of the data as we go beyond the image domain to further illustrate this desired characteristic of our proposed approach}. In fact, the utilized model captures the similarity and dissimilarity between inter-class and intra-class samples without considering domain-specific knowledge.

\subsection{Siamese Architecture}\label{sec:discriminatemodelsub:siamese}

\begin{figure}[ht]
\centering
\includegraphics[width=0.45\textwidth]{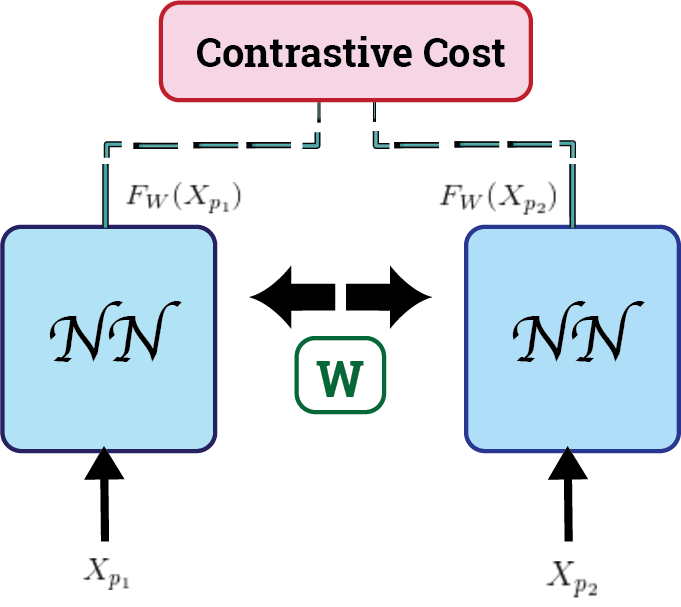}
\caption{Siamese Architecture.}
\label{fig:siamesemodel}
\end{figure}

The discriminative model uses a Siamese architecture~\cite{lecun2005loss,chopra2005learning},~which consists of two identical neural networks.~The aim is to create a target feature subspace for discriminating between similar and dissimilar pairs based on a simple distance metric.
The model is depicted in Fig.~\ref{fig:siamesemodel}.~The general idea is that when two samples belong to a genuine pair~(a pair in which both samples belong to the same category), their distance in the target feature subspace should be as small as possible, while for an impostor pair~(a pair in which the samples belong to different categories), the samples should be as far apart as possible in the output space.~Let $X_{p_{1}}$ and $X_{p_{2}}$ be a pair of samples as the inputs of the system whether in training or testing mode.~The distance between a pair of samples in the target subspace is defined as $D_W(X_{p_{1}},X_{p_{2}})$ (e.g.,~the $\ell_2-norm$ between two vectors) in which $W$ is the parameters of the whole network (weights).
Stated more simply, after training the Siamese architecture, $D_W(X_{p_{1}},X_{p_{2}})$ should be low for genuine pairs and should be high for impostor pairs; this defines the contrastive loss function.
Consider Y as the label which is considered to be $1$ for genuine pairs and $0$ otherwise.
$F$ is the network function which maps the input to the target feature subspace.

The outputs of the Siamese CNNs are denoted by $F_{W}(X_{p_{1}})$ and $F_{W}(X_{p_{2}})$, and both CNNs share the same weights.~The distance is computed as:

\begin{equation}\label{eq1}
D_W (X_{p_{1}},X_{p_{1}}) = {||F_W (X_{p_{1}})-F_W (X_{p_{2}})||}_{2} .
\end{equation}

\subsection{Learning}

\subsubsection{Contrastive Cost}\label{sec:discriminatemodelsub:constrastive}
The goal of the loss function $\mathcal{L}_W(X,Y)$ is to minimize the loss in both scenarios, i.e., of encountering genuine and impostor pairs, so the definition should satisfy two conditions, as follows:
\begin{align}\label{eq2}
\mathcal{L}_W(X,Y) = {\frac{1}{\mathcal{N}}}  \sum_{k=1}^{N} L_W(Y_i,(X_{p_{1}},X_{p_{2}})_i),
\end{align}

\noindent where $\mathcal{N}$ is the number of samples for training and the function $\mathcal{L}_W(Y_i,(X_{p_{1}},X_{p_{2}})_i)$ is defined as below:

\begin{equation} \label{eq3}
\begin{split}
\mathcal{L}_W&(Y_i,(X_{p_{1}},X_{p_{2}})_k) = Y_i*\mathcal{L}_{gen}(D_W(X_{p_{1}},X_{p_{2}})_k)\\ &+ (1-Y_i)*\mathcal{L}_{imp}(D_W(X_{p_{1}},X_{p_{2}})_k,
\end{split}
\end{equation}

where $L_{gen}$ and $L_{imp}$ are defined as follows:

\begin{equation}\label{eq4}
  \begin{cases}
    \mathcal{L}_{gen}(D_W)={\frac{1}{2}}(D_W)^2\\
    \mathcal{L}_{imp}(D_W)={\frac{1}{2}}(\max\{0,M-D_W\})^2.
  \end{cases}
\end{equation}

\noindent where $M$ is a margin which is obtained by cross-validation and $D_W$ stands for $D_W(X_{p_{1}},X_{p_{2}})$.
Moreover, the $max$ argument declares that in case of an impostor pair, if the distance in the target feature space is greater than the threshold $M$,~there would be no loss.

\subsubsection{Input}
For training input, we must have genuine and impostor pairs from the real data.
The goal of training a Siamese architecture is to put genuine and impostor pairs into close and distant manifolds, respectively. To create genuine pairs, we combine samples from the same classes as follows:
$$(X_{p_1},X_{p_2})\ |\  X_{p_1},X_{p_2}\in \mathbf{y}_i$$

On the other hand, to create imposter pairs, we do as follows:
$$(X_{p_1},X_{p_2})\ |\  X_{p_1}\in \mathbf{y}_i,X_{p_2}\in \mathbf{y}_j,i \neq j$$

\subsection{Method Statement}

Assume having an unconditional sample generation setting, with a dataset that is composed of samples $\mathbf{x}^{(i)}$. In unconditional sample generation, the labels $\mathbf{y}^{(i)}$ do not play a role in image generation. However, we need the ground-truth labels to train our Siamese architecture. But, how and why?

As is explained earlier, a Siamese architecture utilizes two identical networks~(networks with the same architectures and parameters~$\theta$) to create a nonlinear mapping from its input domain to a shared Euclidean output feature space:
$$ \psi: \mathcal{X} \rightarrow \mathbb{R}^{m} $$

Due to the weight sharing and contrastive cost, such a scheme guarantees that:

\begin{itemize}
    \item similar samples will stay close in the output feature space;
    \item the model is robust against intra-class samples variations, as the model minimizes intra-class differences;
    \item dissimilar samples will simply be placed in distance places in the output space; and
    \item the model is robust against inter-class samples similarities as the model maximizes inter-class differences.
    
\end{itemize}

Hence, we ideally will have separate clusters, in each of which there are samples from just one particular category. For example, all images belonging to dogs will be placed in one cluster, while pictures of cats will reside in another cluster in the output Euclidean space. \textit{Each cluster is a data manifold that belongs to that class}.

Such a learning paradigm will create a system that is able to:

\begin{enumerate}
    \item recognize the category of a fake sample, i.e., which cluster of data the fake sample belongs to~(determining the closest cluster to classify the fake sample);
    \item determine how close the fake sample is to the cluster or any real data within~(determining the quality of a fake sample given the real samples); and
    \item clarify how diverse the generated fake samples are~(to penalize the model for mode collapse).
\end{enumerate}

This metric will penalize the model for not producing all modes of the data distribution~(comparing with the data inside clusters) as well as classes~(all clusters have some fake samples associated with them). How close a fake sample is to a manifold, determines the \textit{precision}. On the other hand, \textit{recall} refers to how well the generator can produce samples that are similar to the variety of samples in the data manifold~\cite{lucic2018gans}.

\subsection{Siamese Distance Score~(SDS)}

Once the network is trained, we can use it for evaluation purposes. Referring back to Fig.~\ref{fig:siamesemodel}, we technically have one set of weights $\mathbf{W}$ and two copies of one network. Evaluation of \textit{a fake sample} has the following procedure:

\begin{enumerate}
    \item We feed all real samples to the trained neural network and compute the feature vector $F_W(X_{i}^{r})$ for each real sample $X_{i}^{r}$. If we have $N$ real samples,~we will have $N$ feature vectors:
    $$\mathcal{F}_{i}^{r} = F_{W}(X_{i}^{r}), i \in \{1,\ldots,N\}$$
    \item We feed a fake~(synthesized) sample $X_{j}^{s}$ to the network and compute its output feature vector $F_{W}(X_{j}^{s})= \mathcal{F}_{j}^{s}$. If we have $M$ fake samples,~we will have $M$ feature vectors:
    $$\mathcal{F}_{j}^{s} = F_{W}(X_{j}^{s}), j \in \{1,\ldots,M\}$$
    \item We calculate the Euclidean distance of each $\mathcal{F}_{j}^{s}$ with all previously calculated feature vectors $\mathcal{F}_{i}^{r}$~($i \in \{1,\ldots,N\}$), and we pick the $K$ closest ones~($K$ smallest distances $\mathcal{D}_{i}^{j}$ and $j$ is fixed for each fake sample). The index list is denoted with $\mathcal{P}$ for which $\left | \mathcal{P} \right |=K,\mathcal{P} \in \{1,\ldots,N\}$.
    $$\mathcal{D}_{i}^{j}=\left \| \mathcal{F}_{j}^{s} - \mathcal{F}_{i}^{r} \right \|,i \in \{1,\ldots,N\}$$
    \item Among the closest $K$ $\mathcal{D}_{i}^{j}$ distances, we pick their associated $\mathcal{F}_{i}^{r}$.~From that, we extract their associated real samples and labels.
    $$\mathcal{D}^{j}_{\mathcal{P}}\Rightarrow \mathcal{F}_{r}^{\mathcal{P}}\Rightarrow {X}_{r}^{\mathcal{P}},\left | \mathcal{P} \right |=K,\mathcal{P} \in \{1,\ldots,N\}$$
    \item As we have the class of real samples, using a simple majority vote~(K-nearest neighbor algorithm), we determine the class of the fake sample. Basically, the majority vote operates on ${X}_{r}^{\mathcal{P}}$ samples~(~real samples with indexes of $\mathcal{P}$) and their associated classes.
    \item After we determine the class of the fake sample as $C$,~in the cluster of $K$ nearest samples, we take out the real samples that have the determined class label of the fake sample~(~class $C$). The index list of these samples is denoted with $\mathcal{R}$ for which $\left | \mathcal{R} \right |=R,\mathcal{R} \in \{1,\ldots,N\}$.
    $$X_{r}^{\mathcal{R}}, \mathbf{y}(X_{r}^{\mathcal{R}})=C$$
    \item We calculate the distance of the fake sample output feature $F_{s}^{j}$ with all $F_{r}^{\mathcal{R}}$ feature vectors, then compute the average, and denote it as $SDS_{j}$.~The subscript $j$ refers to the fake sample index.~We then average $SDS_{j}$ over all fake samples and call it the \textit{Siamese Distance Score~(SDS)}. 
\end{enumerate}

\section{Experiments}

In order to conduct our experiments, we split our data in three partitions as $\mathcal{D}=\mathcal{G} \cup \mathcal{S} \cup \mathcal{E}$.
$\mathcal{G}$, $\mathcal{S}$, $\mathcal{E}$ will be used for training our generative model, discriminative model, and evaluations, respectively.
It is worth noting that \textbf{(1)} $\mathcal{G} \cap \mathcal{S} \cap \mathcal{E} = \varnothing$,~\textbf{(2)} both $\mathcal{G}$ and $\mathcal{S}$ follow the same data distribution due to the random partitioning, and \textbf{(3)} all class labels in the data are available in $\mathcal{G}$ and $\mathcal{S}$, in balance.
We denote the fake generated data as $\mathcal{F}$.
We run all our experiments 10 times and then report the average.
Such a setup ensures the robustness of the results against possible sensitivity regarding the network weights.
We observed a high stability and very low variance.
This indicates the robustness of the model, relative to small variations of the network weights due to retraining.
This corrects a major drawback of Inception Score~\cite{barratt2018note}.

\subsection{Image Domain}

To ensure a fair set of experiments, we made the following selections for our experiments related to the image domain: We used three datasets that are usually used in the GANs literature: MNIST~\cite{lecun1998gradient},~Fashion-MNIST~\cite{xiao2017fashion}, and CIFAR-10~\cite{krizhevsky2009learning}.
To train our generative model, we used WGAN~\cite{arjovsky2017wasserstein}, due to its stability and robustness to mode collapse~\cite{salimans2016improved}, with DCGAN~\cite{radford2015unsupervised} as the architecture for all datasets.
For all experiments, we fixed the latent dimension~(input noise $z$) to 100 and the noise distribution to be $\mathcal{N}(0,1)$. The batch size is set to 64.
We train our GAN for 20, 20,~and 100 epochs for MNIST, Fashion-MNIST, and CIFAR-10,~respectively.

To train our discriminative Siamese network, we used a three-layer convolutional neural network, followed by a fully-connected layer of size 1024 as the output embedding space. Furthermore, we used batch normalization~\cite{ioffe2015batch} to improve the training.
We utilized LeakyReLU~\cite{maas2013rectifier} for layers' activation to avoid encountering dead ReLUs and zero gradients.
In this work, to calculate SDS, we used a simple nearest neighbor (K-nearest neighbor in which $K=1$) classifier as we did not observe a unique difference by picking $K\neq1$.

\subsubsection{\textbf{Quality}}

\begin{figure}[ht]
\centering
\includegraphics[width=0.45\textwidth]{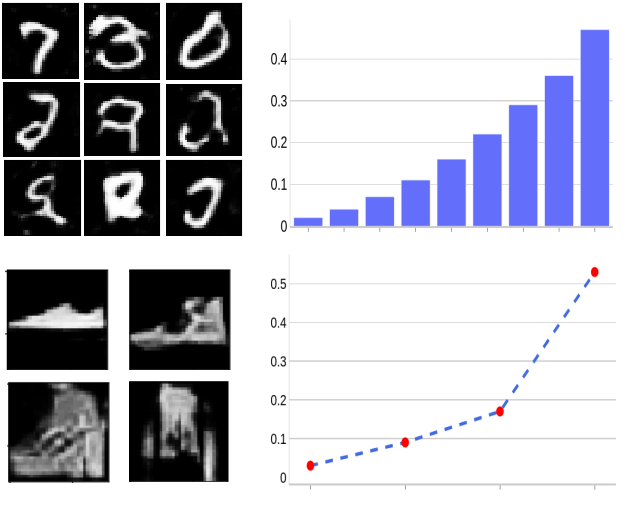}
\caption{Demonstration of the quality measurement of fake samples which is consistent with human evaluation.~The horizontal axis is the index of images from left to right and top to bottom.~The vertical axis is the SDS score of the sample.}
\label{fig:quality}
\end{figure}

To assess the visual quality, we demonstrate the evaluation of fake samples on MNIST and Fashion-MNIST in Figure~\ref{fig:quality}.
As can be observed,~as the SDS increases,~the visual image quality decreases.
Such observation is aligned with human evaluation.

\subsubsection{\textbf{Mode dropping and invention}}

To evaluate \textbf{mode dropping}~(not generating a class of data) and \textbf{mode invention}~(generating samples that do not belong to any class), we used MNIST,~Fashion-MNIST, and CIFAR-10 datasets by fixing all their images' sizes to $32 \times 32$. Each dataset has 10 classes. Here's how we set up our experiments:
\begin{itemize}
    \item We pick $\mathcal{S}$ so it only has 5 classes and train our Siamese model with it.
    \item We create a test set $\mathcal{T}$ from $\mathcal{E}$ which only has $i$ classes.
    \item For each $i$ we measure SDS.

\end{itemize}

\begin{figure}[ht]
\centering
\includegraphics[width=0.45\textwidth]{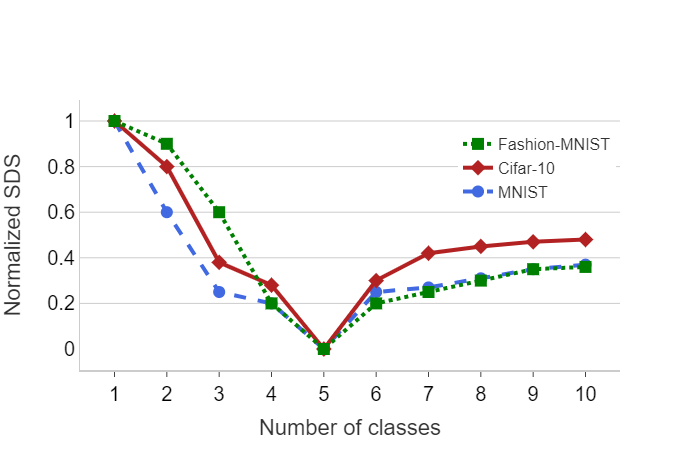}
\caption{SDS is sensitive to mode dropping and mode invention. The higher the SDS is, the worst the results are.~The score for each dataset has its own scale. For each dataset experiment, its scores are normalized according to its own max-min obtained scores.}
\label{fig:modeeval}
\end{figure}

The results are depicted in Figure.~\ref{fig:modeeval}.~As can be observed, increasing the number of classes from 1 to 5, gives a better score, as fewer classes mean mode dropping.~Also, by increasing the number of classes beyond 5,~SDS increased as well, which shows the detection of mode invention.

\begin{figure}[ht]
\centering
\includegraphics[width=0.45\textwidth]{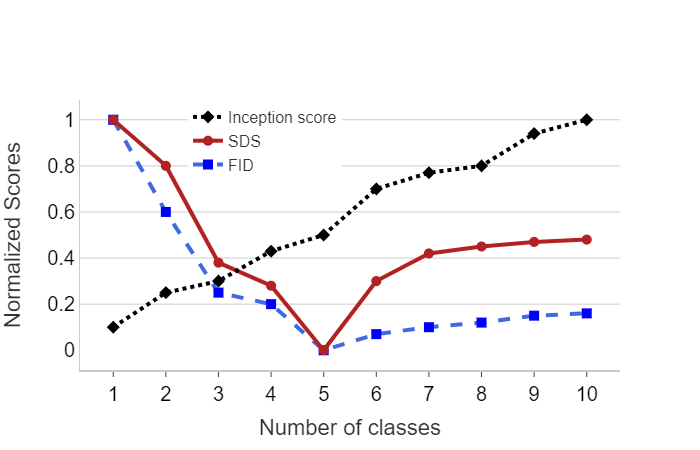}
\caption{Comparison of \textit{Inception Score}, \textit{FID},~and our proposed metric~\textit{SDS}.~The score for each dataset has its own scale. For each dataset experiment, its scores are normalized according to its own max-min obtained scores.}
\label{fig:modecompare}
\end{figure}

Figure~\ref{fig:modecompare} demonstrates the comparison of our method SDS with the other two popular methods in the literature. SDS and FID are both good in capturing mode dropping. However, SDS shows higher sensitivity in mode dropping.

\subsubsection{\textbf{Intra-class mode collapse}}

As we mentioned earlier, the intra-class mode collapse~(mode dropping) refers to the situation that GAN can generate all modes~(classes) of data but only one or few examples from each mode~(fails to generate a variety within class samples).~To assess SDS metric robustness to this phenomena, we conduct the following experiments:

\begin{itemize}
    \item We pick $\mathcal{S}$ so it has all classes and trains our Siamese model with it.
    \item We create a test set $\mathcal{T}$ from $\mathcal{E}$ which also has all classes. However, only $p$ percentage of each class's samples are used~(note that $\mathcal{S}$ and $\mathcal{T}$ are mutually exclusive and $|\mathcal{T}|=|\mathcal{E}| \times p$).
    \item For each class, regardless of $p$, we select an identical number of samples~(say $\mathcal{K}$ number of samples per class).~Note that $\mathcal{K}$ is maximum when $p=1$.~We select $\mathcal{K}$ as $\mathcal{K}=|\mathcal{T}| / \mathcal{C}$ in which $\mathcal{C}$ is the number of classes.
    \item For each $p$ we measure SDS and report the results.
\end{itemize}

\begin{figure}[ht]
\centering
\includegraphics[width=0.45\textwidth]{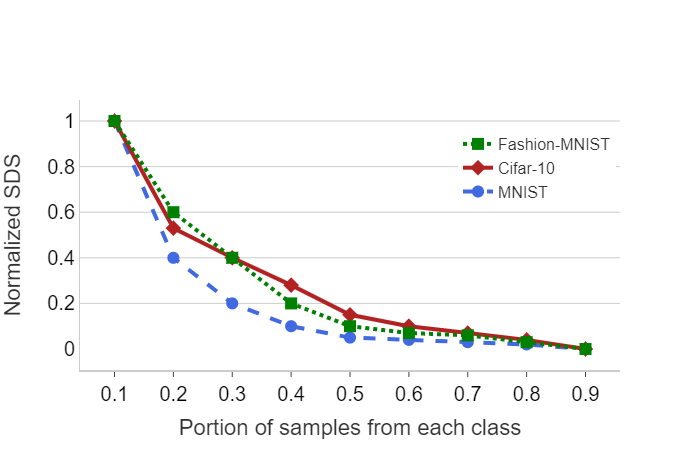}
\caption{SDS is sensitive to intra-class mode dropping.~The higher the SDS is, the worse the results are.~The score for each dataset has its own scale. For each dataset experiment, its scores are normalized according to its own max-min obtained scores.}
\label{fig:intraclass}
\end{figure}

The results are depicted in Figure~\ref{fig:intraclass}.~The metric is sensitive to mode dropping. Based on the evaluation approach, the nearest neighbor algorithm is forced to pick a sample as the closest one. If there are only a few modes available, the algorithm picks from the samples that do not actually represent the variety of the samples available in $\mathcal{T}$. Henceforth,~such behavior was expected.~This is one of the desired characteristics of the proposed approach.

\subsection{GANs Models Evaluation}

To investigate the feasibility of using our metric for different GANs models' comparison,~we shift our focus towards the model evaluation aspects of the work.~For the baseline, we used the GAN with the non-saturating update rule~(NS) as proposed by Goodfellow~\cite{goodfellow2014generative}.

We analyze three different generally utilized GAN models on the Cifar10 dataset.
\textbf{(1)} Baseline GAN~\cite{goodfellow2014generative},
\textbf{(2)} GAN with spectral normalization~(SN)~\cite{miyato2018spectral},~and
\textbf{(3)} WGAN with gradient penalty~(GP)~\cite{gulrajani2017improved}. The utilized architecture is DCGAN.
The results are given in Table~\ref{tab:gancompare}.

\begin{table}[ht]
\caption{The comparison of different GANs models using the FID and SDS metrics.}
\label{tab:gancompare}
\centering
 \begin{tabular}{||c c c||} 
 \hline
 \  & FID & SDS \\ [0.5ex] 
 \hline\hline
 Baseline & 53.72 $\pm$ 5.43 & 0.53 $\pm$ 0.08 \\ 
 \hline
 WGAN + GP & 38.48 $\pm$ 2.73 & 0.21 $\pm$ 0.04 \\ 
 \hline
 SN & 31.25 $\pm$ 2.17 & 0.15 $\pm$ 0.03 \\ [1ex]
 \hline
 \end{tabular}
\end{table}

In concurrence with the results obtained in \cite{miyato2018spectral} to rank the models, SDS also reports the same GANs ranking based on their generated samples quality as well as being consistent with FID.
We computed both FID and SDS scores from 10,000 real and generated samples.


\subsection{Utilization of the Approach Beyond the Image Domain}

To assess the generalizability of our proposed metric, we considered the healthcare domain and the Electronic Health Records~(EHRs) data, which have very different statistics and characteristics as compared to image data.

We performed our experiments with the UCI Epileptic Seizure Recognition
dataset~\cite{andrzejak2001indications}.~This dataset classifies brain activities; the core task is brain seizure classification.
Approximately 20\% of the samples are classified as seizure activity~(we have only two class labels).
The number of features for each sample is 179 and there is a total of 11500 samples.
The first 178 features are associated with the Electroencephalogram~(EEG) values, while the last one is the class label.

Although the main goal of this work is to evaluate GANs,~our proposed approach can be extended to evaluate any kind of generative model. To showcase that, we picked three different successful generative models in the healthcare domain -- \textit{Variational Autoencoders~(VAEs)}~\cite{kingma2013auto},
\textit{medGAN}~\cite{choi2017generating}, and
\textit{corGAN}~\cite{torfi2020cor} -- to generate the synthetic one-dimensional data: 

\begin{itemize}

\item \textbf{Variational Autoencoder (VAE):} For both the encoder and the decoder, a 1D convolutional neural network is utilized with two hidden layers of 128.

\item \textbf{medGan:} The medGAN \cite{choi2017generating} architecture includes:
\textbf{(1)} fully-connected layers,
\textbf{(2)} shortcut connections to augment the generator, and
\textbf{(3)} minibatch-averaging~\cite{choi2017generating} to overcome mode collapse.
Although medGAN has been originally proposed for generating discrete records,~we removed its autoencoder part to directly generate continuous records.

\item \textbf{corGAN:} The corGAN \cite{torfi2020cor} architecture consists of the following elements: 
\textbf{(1)} 1-D convolutional neural network for discriminator and generator,
\textbf{(2)} WGAN training regime.

Both medGAN and corGAN have proven to be successful methods in generating discrete and continuous healthcare records.

\end{itemize}

\begin{figure}[ht]
\centering
\includegraphics[width=0.45\textwidth]{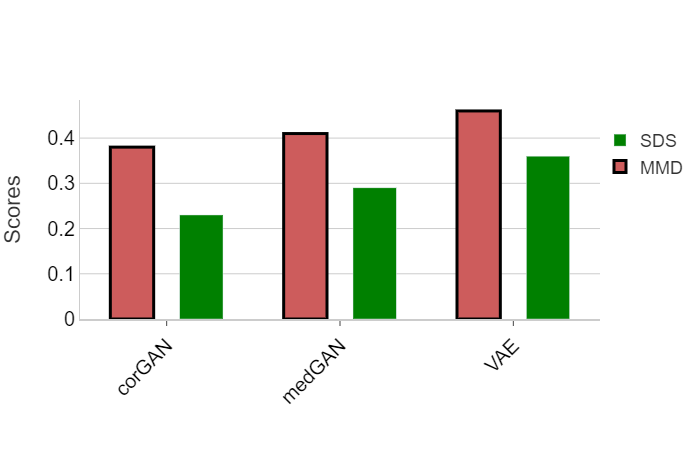}
\caption{The comparison between different generative models using MMD and SDS.
As can be seen,~our proposed metric can effectively rank generative models concur with the MMD score.}
\label{fig:mmdsds}
\end{figure}

After training our Siamese architecture with 3 fully-connected layers, each of size 128, we evaluate the averaged SDS score.
We report the results of our score along with Maximum Mean Discrepancy~(MMD)~\cite{gretton2007kernel}, which is known to be an effective sample-based GANs evaluation metric~\cite{xu2018empirical}.
For both MMD and SDS, an equal number of synthetic and real samples are chosen for calculation.~The results are depicted in Figure~\ref{fig:mmdsds}. As can be observed, SDS is consistent with MMD, regarding comparison of the three models.

\section{Conclusion}

We proposed an evaluation metric for GANs which relies on Siamese neural networks.
Our metric can be applied in evaluation of any generative model.
The proposed approach enables us to evaluate GANs without the need for any pretrained classifiers.
We empirically proved that our method can address different GANs failure situations~(mode dropping,~mode invention,~intra-mode collapse) and is sensitive to visual quality aligned with human evaluation.
Finally,~the significant advantage of the proposed approach is its domain agnostic characteristics that can be utilized beyond the image domain and in a variety of applications.








%
\bibliography{main}
\bibliographystyle{unsrt}

\end{document}